\documentclass[conference]{IEEEtran}
\IEEEoverridecommandlockouts
\usepackage{cite}
\usepackage{amsmath,amssymb,amsfonts}
\usepackage{algorithmic}
\usepackage{graphicx}
\usepackage{textcomp}
\usepackage{xcolor}
\def\BibTeX{{\rm B\kern-.05em{\sc i\kern-.025em b}\kern-.08em
    T\kern-.1667em\lower.7ex\hbox{E}\kern-.125emX}}
\begin{document}

\title{A Comparative Analysis Between the Additive and the Multiplicative Extended Kalman Filter for Satellite Attitude Determination}

\author{
\IEEEauthorblockN{1\textsuperscript{st} Hamza A. Hassan\footnote{Emails for the author: hyusuf19,wtolst19,rl52dw,ikizil22\@student.aau.dk}}
\IEEEauthorblockA{\textit{Dept. of Electronic Systems} \\
\textit{Aalborg University}}
\and
\IEEEauthorblockN{2\textsuperscript{nd} William Tolstrup}
\IEEEauthorblockA{\textit{Dept. of Electronic Systems} \\
\textit{Aalborg University}}
\and
\IEEEauthorblockN{3\textsuperscript{rd} Johanes P. Suriana}
\IEEEauthorblockA{\textit{Dept. of Electronic Systems} \\
\textit{Aalborg University}}
\and
\IEEEauthorblockN{4\textsuperscript{th} Ibrahim D. Kiziloklu}
\IEEEauthorblockA{\textit{Dept. of Electronic Systems} \\
\textit{Aalborg University}}
}

\maketitle

\begin{abstract}
\footnote{Emails for the authors: hyusuf19,wtolst19,jsuria22,ikizil22@student.aau.dk}The general consensus is that the Multiplicative Extended Kalman Filter (MEKF) is superior to the Additive Extended Kalman Filter (AEKF) based on a wealth of theoretical evidence. This paper deals with a practical comparison between the two filters in simulation with the goal of verifying if the previous theoretical foundations are true.
The AEKF and MEKF are two variants of the Extended Kalman Filter that differ in their approach to linearizing the system dynamics. The AEKF uses an additive correction term to update the state estimate, while the MEKF uses a multiplicative correction term. The two also differ in the state of which they use. The AEKF uses the quaternion as its state while the MEKF uses the Gibbs vector as its state.
The results show that the MEKF consistently outperforms the AEKF in terms of estimation accuracy with lower uncertainty. The AEKF is more computationally efficient, but the difference is so low that it is almost negligible and it has no effect on a real-time application.
Overall, the results suggest that the MEKF is a better choise for satellite attitude estimation due to its superior estimation accuracy and lower uncertainty, which agrees with the statements from previous work.
\end{abstract}

\begin{IEEEkeywords}
Multiplicative, Additive, Extended Kalman Filter, Satellite Attitude Determination, Quaternion representation 
\end{IEEEkeywords}

\section{Introduction}
In satellite attitude determination systems the question arises of which state to use to determine the current attitude system of the SO3 group, the group of all rotation matrices in 3D. The attitude matrix is a nine-component matrix with six constraints (three norm constraints and three orthogonality constraints) and is not a common way of parameterizing the attitude state. Roll, pitch, and yaw values are more common as they directly represent the SO3, however, as commonly known, they suffer from gimbal lock, discontinuities, and non-commutativity in terms of computing the attitude matrix. The quaternion is the lowest dimensional non-singular representation of the SO3 consisting of four components constrained to have the norm 1. This paper investigates how the quaternion is used in the derivatives of the Kalman Filter (KF) to find the optimal estimation given the current measurements. The focus will be on the Additive Extended Kalman Filter (AEKF) and the Multiplicative Extended Kalman Filter (MEKF). 

MEKF was first used in 1969 \cite{b13}\cite{b14} and used in several NASA multimission satellites\cite{b4} and is considered a standard attitude estimation algorithm. The MEKF uses a three-dimensional vector representation of the error to update a quaternion reference to estimate the optimal quaternion estimation and the output of the MEKF is guaranteed to be a unit quaternion. The AEKF has the state of the four-dimensional quaternion and considers the quaternion components as four distinct components and uses the standard Extended Kalman Filter (EKF) equations to update the state and then uses brute force normalization. 

In the theoretical discussions of the MEKF and the AEKF, the MEKF has been considered superior by most authors. A discussion on the singularity of the covariance matrix or its ill-conditioning in the AEKF has been made, but there are methods to ensure the stability of the AEKF. Nevertheless, the MEKF is still preferred \cite{b5}. Furthermore, in the theoretical discussions of the MEKF vs. the AEKF, it has been noted that numerical issues regarding the AEKF might arise. In practice, however, this has caused no issue \cite{b6}. Markley \cite{b3} shows that the MEKF has the lowest computational time because of its three-dimensionality, and since the AEKF rests on a less secure theoretical foundation, he concludes that there is no valid reason for using the AEKF over the MEKF.

As such, there have been many theoretical comparisons between the AEKF and the MEKF but none of these have been practical and therefore a comparison between the two algorithms is sought after. There have been some comparisons done such as the one by Zhang et al. \cite{b7} which compared many filtering algorithms with the MEKF but not with the AEKF. Zamani et al. \cite{b12} compared many state-of-the-art estimation algorithms of which the MEKF was one of them but the AEKF was omited due to its apparent shortcomings. Markley \cite{b18} showed that  MEKF and AKEF were compared and derived that the covariance matrix of the AEKF was not singular and stated that the AEKF brute force normalization was optimal under standard linear filtering assumptions, however, concluded that because most realistic models are non-linear there are still no valid reasons to prefer AEKF to the MEKF. 

Thus as can be seen there have been many discussions on the conditioning, and singularity of the covariance but no practical, in-depth comparison between the AEKF and MEKF has been given despite the abundance of the theoretical foundation for the MEKF being superior to the AEKF. Therefore, it is the aim of this paper to compare the AEKF and the MEKF in simulation on parameters such as the computational time, conditioning of the covariance matrix, the norm of the covariance matrix, and comparing errors in the estimation. The standard MEKF has a gyroscope bias estimator, but this will be disregarded in the comparison as it is not relevant to this discussion.

This paper is structured as follows:
\ref{Sec:methods} describes the methods used. \ref{Sec:filters} gives an overview of some of the derivations of KF. \ref{Sec:simulation} shows the implementation in Simulink, and lastly \ref{Sec:results} shows the findings and gives a concise comparison.

\section{Methods}\label{Sec:methods}

\subsection{Rotations}
Since rotations play a significant role in the analysis of AEKF/MEKF, it is necessary to discuss them in detail, and for this, a more in-depth discussion can be read in the introduction given by Markley \cite{b2}.
Due to the limited number of rotational planes, rotations in two dimensions are straightforward.
The center of rotation is the only parameter that is freely adjustable.
Thus, a single angle can be used to represent an object's orientation in 2D, although discontinuities occur which can be solved by introducing a 2D-heading vector constrained to be 1. The same method is used in 3D. \\
In 3D, rotations are substantially more challenging. An axis of rotation, that is denoted by a vector, now serves as the origin of rotation with the plane of rotation being perpendicular.
The drawback of this method is a condition known as \textit{Gimbal Lock}. It refers to the loss of a degree of freedom when two of the three gimbal axes are forced into a parallel configuration, collapsing the two rotations into a single one, making it impossible to rotate out of it.

Instead, the quaternion is the representation of choice because it is the lowest-dimensional parameterization free of singularities, such as a gimbal lock. A quaternion \textbf{q} has a scalar part \textit{w} and a three-vector part $\textbf{q}_{1:3}$

\begin{equation}\label{eq:def:quat}
    \textbf{q} = \begin{bmatrix}
        \textit{w}\\
        \textbf{q}_{1:3}
    \end{bmatrix}  
    \textnormal{  where       }
    \textbf{q}_{1:3} = \begin{bmatrix}
        q_1\\
        q_2\\
        q_3
    \end{bmatrix}
\end{equation}

The quaternion, however, must obey its normalizing constraint that is
\begin{equation}\label{eq:def:quatnorm}
    ||\textbf{q}|| = \sqrt{w^2+q_1^2+q_2^2+q_3^2} = 1
\end{equation}

Rodrigues parameters is an effective algorithm for rotating a vector in space given an axis and angle of rotation in three-dimensional rotation theory.
By extension, this can be used to convert each of the three basis vectors in order to calculate a rotation matrix in SO3.
They were later used by J. Willard Gibbs, who invented modern vector notation, therefore the vector of Rodrigues parameters is referred to as the \textit{Gibbs vector}. It relates to the quaternion as 

\begin{equation}\label{eq:def:gibbs}
    \textbf{g} = \frac{\textbf{q}_{1:3}}{w}
\end{equation}

The Rodrigues parameters provide a 2:1 mapping because both \textbf{q} and $-\textbf{q}$ represent the same attitude matrix. A full explanation can be found in \cite{b1}. The consequence of this is that at $180^{\circ}$, the Gibbs vector becomes infinite. It therefore only excels when representing small angles.

The kinematic equation for the quaternion as noted by \cite{b2} is given as

\begin{equation}\label{eq:quatkinematics}
    \Dot{\textbf{q}} = \frac{1}{2}\begin{bmatrix}
        \omega\\
        0
    \end{bmatrix}
    \otimes \textbf{q}
\end{equation}

where $\omega$ is the angular velocity vector. It is important to note that Eq. (\ref{eq:quatkinematics}) preserves the normalization constraint of \textbf{q}. If the norm constraint is violated, it can easily be corrected by dividing \textbf{q} by its 2-norm.

\subsection{Star Tracker}\label{sec:startracker}
A star tracker is a digital camera with a specialized lens at the focal plane. Star trackers are very precise sensors that can detect a star vector within the body frame of the satellite and by virtue of the fact that the location of the stars is known in the frame of the Earth, it is possible to determine the current attitude of the satellite. For attitude determination, a single star tracker can be an under-determined observer if only a single star is visible. Having multiple star trackers is an advantage in that it can improve the robustness of the orientation estimate to noise and is redundant in terms of malfunctions in one of the other star trackers. Furthermore, with enhanced coverage, it can increase the accuracy.
They typically operate at a fairly low frequency between $0.5$ and $10$ Hz, which is considerably slower than other sensors found on a satellite, e.g. a gyroscope \cite{b1}.
A disadvantage of star trackers is that they cannot be used when pointed towards the sun which means that on low orbit satellites that are only in darkness 30\% of the time, they need multiple star trackers so that at least one is not pointing towards the sun \cite{b15}.

When the satellite has been deployed it has to detect the stars without knowing its initial orientation. This is called lost in space mode. When an initial attitude estimation has been determined, the satellite switches to a recursive estimator that can run at much higher frequencies \cite{b16}.

Since the star tracker has a specific field of view, $\theta_{FoV}$, it can be determined if the star is within this field of view, if the star vector and $b=[0,0,f]$ have an angle smaller than 

\begin{equation}\label{eq:starangle}
    \cos(\theta_{FoV})<  \frac{(\vec{a}\bullet \vec{b})}{|\vec{a}||\vec{b}|} \frac{1}{2}
\end{equation}

\subsection{Wabha problem and Davenport solution} 
Given a series of vectors in different coordinate representations, i.e the body- and inertial coordinate representations, \textbf{b} and \textbf{r} from a star camera, the question of how to find the attitude matrix that aligns the two frames, i.e $\textbf{Ar} = \textbf{b}$ arises.
The problem is to find the orthogonal matrix \textbf{A} that solves this. The methods for solving it are called \textit{single frame} methods as they do not take into account the system dynamics as opposed to filtering methods such as the KF and its derivatives.\cite{b8}

\begin{equation}\label{eq:Wahba}
    \textbf{A}=\operatorname*{arg\space min}_{ \left\| \textbf{A}\right\| =1} \sum_{i=1}^{N}w_i\left\| \textbf{b}_i-\textbf{A}\textbf{r}_i \right\|^2
\end{equation}

There are many propositions for solutions for this problem, but the most robust methods is the \textit{q-method} developed by Davenport which solves the optimization problem.
Following the derivations as from \cite{b8, b1}, it ends with the following
\begin{equation} 
\label{eq:kfinal}
\begin{split}
    0=\lambda(\textbf{q})-2\textbf{K}\textbf{q}\\
        \textbf{K}\textbf{q}=\lambda (\textbf{q})
\end{split}
\end{equation}

Where \textbf{K} is the Davenport matrix and $\lambda$ denoted the eigenvector.

It can be seen clearly in Eq. (\ref{eq:kfinal}) that the quaternion estimate that solves the optimization problem is the eigenvector of the Davenport matrix which has the largest eigenvalue. Thus, to find the quaternion measurement from a set of vectors in the body frame and a set of vectors in the inertial frame, the Davenport matrix is defined and the eigenvector with the largest eigenvalue is the quaternion estimate.
\section{Filters}\label{Sec:filters}

\subsection{Kalman Filter}
The KF \cite{b19} is an effective recursive filter that determines the internal state of a linear dynamic system from a series of noisy measurements.
The linear-quadratic-Gaussian (LQG) control problem is solved by the KF.

The internal state is typically substantially larger (has more degrees of freedom) than the few "observable" characteristics that are monitored in the majority of applications.
However, the KF can estimate the full internal state by combining different measurements. In other words, it aims to find the estimated state on the basis of the previous state, a measurement and some Kalman gain. 

The KF can be summarised in a single equation

\begin{equation}\label{eq:KF}
 \hat{\textbf{x}}_{k|k} = \hat{\textbf{x}}_{k|k-1} + \textbf{K}_k\textbf{y}_k
\end{equation}

With $\hat{\textbf{x}}_{k|k}$ being the estimated state that is sought after. $\textbf{K}_k$ is the Kalman gain, $\textbf{y}_k$ is the measurement residual which is the error between the measurement and the expected measurement, and $\hat{\textbf{x}}_{k|k-1}$ is the previous state.
The Kalman gain is used in the KF to determine how much weight to give to each measurement in order to produce the optimal estimate of the system's state. It is used to update the state estimate at each time step, and it determines the relative importance of the different measurements. The filter can weigh more recent measurements more with a higher gain.

Aside from the estimated state, the estimate covariance is also sought after, as it is a measure of the estimated accuracy of the estimated state.
The filter follows a two step method: A \textbf{prediction} step and an \textbf{update} step. The two phases typically alternate, with the prediction progressing the state until the next planned observation and the update incorporating the observation. It is not necessary, however, as multiple measurement updates may occur between prediction steps. But in the most common implementations, the prediction occurs multiple times before the update. In this case, the prediction is based on the integrated $\omega$ from the gyroscope and the update happens from measurements from the much slower star tracker.

\subsection{Extended Kalman Filter}
While the simple KF only works for a linear function, the EKF \cite{b20} works for any non-linear functions as long as they are differentiable. This is much more useful, as most problems in the real world are non-linear.
The problem with this is that the KF works with Gaussian distributions. If a gaussian is fed a linear function, the output is still gaussian. This is not the case for non-linear functions where the output is non-gaussian.
The solution is to approximate the linearity of the non-linear functions and this is done by using Taylor Series.
Like the simple KF, it consists of a \textbf{prediction} and \textbf{update} step. 
\vspace{2mm}
\hrule
\vspace{2mm}
\textbf{Prediction}
\begin{gather}
    \hat{\textbf{x}}_{k|k-1} = f(\hat{\textbf{x}}_{k-1|k-1}, \textbf{u}_k)\\\nonumber
    \textbf{P}_{k|k-1} = \textbf{F}_k\textbf{P}_{k-1|k-1}\textbf{F}_k^T+\textbf{Q}_k
\end{gather}

\hrule
\vspace{2mm}
\textbf{Update}
\begin{gather}
    \textbf{y}_k = \textbf{z}_k - h(\hat{\textbf{x}}_{k|k+1})\\\nonumber
    \textbf{S}_k = \textbf{H}_k\textbf{P}_{k|k-1}\textbf{H}_k^T+\textbf{R}_k\\\nonumber
    \textbf{K}_k = \textbf{P}_{k|k-1}\textbf{H}_k^T\textbf{S}_k^{-1}\\\nonumber
    \hat{\textbf{x}}_{k|k} = \hat{\textbf{x}}_{k|k-1} + \textbf{K}_k\textbf{y}_k\\\nonumber
    \textbf{P}_{k|k} = (\textbf{I}-\textbf{K}_k\textbf{H}_k)\textbf{P}_{k|k-1}
\end{gather}

\hrule
\vspace{2mm}

Again, the goal is to find the estimated state $\hat{\textbf{x}}_{k|k}$ and the estimate covariance $\textbf{P}_{k|k}$. 
The state-transition $\textbf{F}_k$ and observation $\textbf{H}_k$ matrices are defined as follows
\begin{equation}
    \textbf{F}_k = \frac{\delta f}{\delta\textbf{x}}\\
    \textnormal{ and }
    \textbf{H}_k = \frac{\delta h}{\delta\textbf{x}}
\end{equation}

Where the jacobian is used to transform from the non-linear space into a linear one. The state-transition is used to predict the state of a dynamic system at the next time step based on its current state, and the observation matrix is used to map the state to the measurement space.
The matrices $\textbf{Q}_k$ and $\textbf{R}_k$ are the covariance matrices for process- and measurement noise respectively. These parameters can be tuned, and  are used to model the uncertainty in the system's state and measurements, respectively.

\subsection{Additive Extended Kalman Filter}
The AEKF considers each element of the four-component quaternion to be independent and does not take extra consideration to the quaternion norm. In the update step, the quaternion is updated with measurement by stating $\hat{\textbf{q}}_{k|k}=\textbf{q}_{k|k-1}+\textbf{K}_k\textbf{y}_k$ and afterward the norm is normalized by $\frac{\textbf{q}}{||\textbf{q}||}$ by assuming a deterministic relationship between the quaternion. Thus the AEKF is simply the EKF but normalizes the quaternion norm in each step. Furthermore, the normalization can be considered a measurement update where it is "observed" the quaternion has the norm one. 

As expected and as stated before, several shortcomings exist with this approach. Consider that the AEKF estimation is unbiased and therefore the expectation can be given as $E\{\hat{\textbf{q}}\}=\textbf{q}^{true}$ and the additive quaternion error can be defined as 

\begin{equation}
\label{eq:errorquat}
    \Delta \textbf{q} \equiv \textbf{q}^{true}-\hat{\textbf{q}}
\end{equation}
This shows that by definition the quaternion estimate must lie outside the quaternion hyperplane. Furthermore, discussions on the covariance matrix have shown that it becomes ill-conditioned. \cite{b1}

\begin{equation}
   \left\| \hat{\textbf{q}}  \right\|^2=\left\| \textbf{q}^{true}-\Delta \textbf{q}  \right\|^2 = \left\| \textbf{q}^{true}  \right\|^2 -  2\Delta \textbf{q}^T\textbf{q}^{true}+ \left\|\Delta \textbf{q}\right\|^2
\end{equation}

\begin{equation}
E\{\| \textbf{q} \|^2 \}=1+E\{\| \Delta \textbf{q} \|^2 \}
\end{equation}

\begin{figure}[htbp]
    \centering
    \includegraphics[width = 0.3\textwidth]{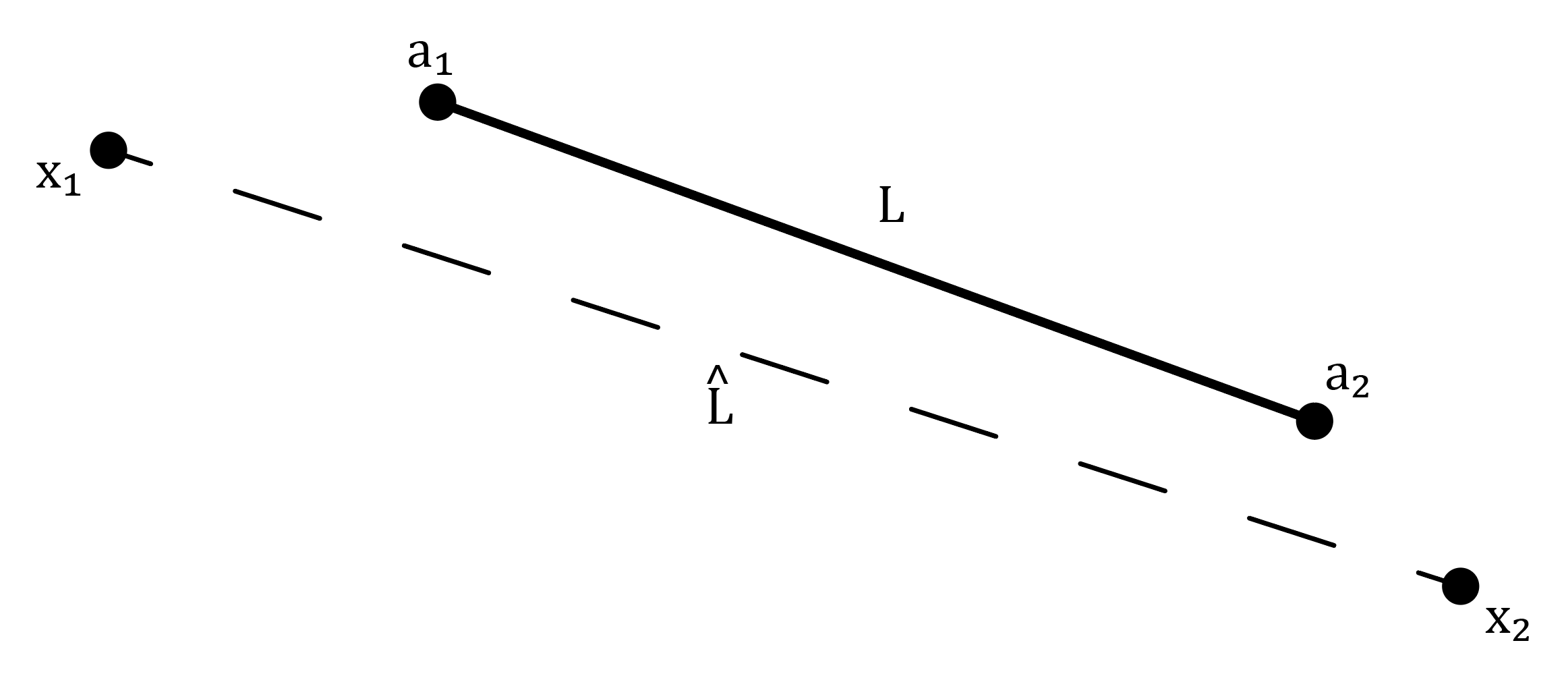}
\caption{Estimating the vertices of a line. The deterministic relationship of the length does not hold in this situation.}
\label{fig:estimation}
\end{figure}

Furthermore, the deterministic assumption is false. Consider Fig. \ref{fig:estimation} where there is a rod that has a length $L$ and turns with a certain angular velocity $\omega$. It has endpoints in $a_1$ and $a_2$. Consider now the initial estimate of $x_1$ and $x_2$ at the first time step. As time goes on, the estimates will get infinitesimally closer to the actual vertices of the rod, but the assumption that the length can be deterministically found from the estimated points is false as can be seen from the figure, as the estimated length is different from the true one.
The ramification of this is that while there is a deterministic relationship between the quaternion elements $\textbf{q}_1,\textbf{q}_2,\textbf{q}_3$ and $\textbf{q}_4$ it does not mean that this deterministic relationship has a correspondence when talking about their estimates $\hat{\textbf{q}}_1,\hat{\textbf{q}}_2,\hat{\textbf{q}}_3$ and $\hat{\textbf{q}}_4$. This illustrates why one cannot "simply" normalize the quaternion as done in AEKF as it assumes a deterministic relationship between the estimates when such a relationship is not known. \cite{b17}

\subsection{Multiplicative Extended Kalman Filter}
The MEKF uses the quaternion as the actual attitude representation and a three-component state vector $\delta \textbf{a}_g$ for the representation of attitude errors, here being the Gibbs vector. As opposed to taking the sum of the estimated quaternion and the error quaternion as in the AEKF, in the MEKF the state is Gibbs projection of the product of the estimate and the error quaternion.

\begin{equation}\label{eq:def:qpred}
    \textbf{q}_p = \delta \textbf{q}(\textbf{a}_{g}) \otimes \hat{\textbf{q}}^+
\end{equation}

The Gibbs vector is used here as the error vector, but many different error representations can be used, cf. \cite{b2}.

The four-component correctly normalized $\hat{\textbf{q}}^+$ is not part of the EKF, but the reset operation moves the a posteriori estimate, i.e, $\hat{\textbf{q}}^+$ into this variable, keeping the error quaternion small.\\
Following the procedure for finding the matrices \textbf{Q} and \textbf{H} from \cite{b1}, and \textbf{G} and \textbf{F} from \cite{b2}, it is found that

\begin{gather}\nonumber
    \textbf{Q} = 
    \begin{bmatrix}
        (\sigma_v^2 \Delta t + \frac{1}{3} \sigma_u^2 \Delta t^3)I_{3\times3} & -(\frac{1}{2}\sigma_u^2 \Delta t^2)I_{3\times3}\\
        -(\frac{1}{2}\Sigma_u^2 \Delta t^2)I_{3\times3} & (\sigma_u^2 \Delta t)I_{3\times3}
    \end{bmatrix}\\\nonumber
    \textbf{H} =
    \begin{bmatrix}
        I_{3\times3} & 0_{3\times3}
    \end{bmatrix}\\\label{eq:MEKF:Matrices}
    \textbf{G} = 
    \begin{bmatrix}
        -I_{3\times3} & 0_{3\times3}\\
        0_{3\times3} & I_{3\times3}
    \end{bmatrix}\\\nonumber
    \textbf{F} =
    \begin{bmatrix}
        -[\omega \times] & -I_{3\times3}\\
        0_{3\times3} & 0_{3\times3}
    \end{bmatrix}
\end{gather}

The identity component in the sensitivity matrix \textbf{H} in Eq. (\ref{eq:MEKF:Matrices}) is possible because of the emulated star tracker output. The quaternion output is converted to a Gibbs vector which is then used in the filters, which gives an identical mapping.
$[\omega\times]$ in \textbf{F} denotes the cross product matrix of the angular velocity $\omega$

\vspace{2mm}
\hrule
\vspace{2mm}
\textbf{Prediction Step}
\begin{gather}\label{eq:MEKF:PredictionStep}
    \hat{\textbf{q}}_p = \frac{1}{2}\Xi \hat{\textbf{q}}^+_k\\\nonumber
    \hat{\textbf{P}} = \textbf{FPF}^T+\textbf{GQG}^T
\end{gather}

\hrule
\vspace{2mm}
\textbf{Update Step}
\begin{gather}\nonumber
    \textbf{S} = \textbf{H}\hat{\textbf{P}}\textbf{H}^T+\textbf{R}\\\nonumber
    \textbf{K} = \hat{\textbf{P}}\textbf{H}^T\textbf{S}^{-1}\\\nonumber
    \textbf{P} = \hat{\textbf{P}} - \hat{\textbf{P}}\textbf{KH}\\\nonumber
    \textbf{q}_e = \textbf{y} \otimes \hat{\textbf{q}}_p^{-1}\\\label{eq:MEKF:UpdateStep}
    \textbf{a}_g = \frac{\textbf{q}_{e1:3}}{w}\\\nonumber
    \Delta\hat{\textbf{x}}^+ = \textbf{K}\textbf{a}_g\\\nonumber
    \delta\textbf{q}(\textbf{a}_g) = \begin{bmatrix}
        \Delta\hat{\textbf{x}}^+_{1:3}\\
        2
    \end{bmatrix}\\\nonumber
    \hat{\textbf{q}}_p^{unnorm} = \hat{\textbf{q}}_p^- + \frac{1}{2} \Xi (\delta\textbf{q}(\textbf{a}_g) \otimes \hat{\textbf{q}}^+)\\\nonumber
    \hat{\textbf{q}}_p^+ = \hat{\textbf{q}}_p^{unnorm} / ||\hat{\textbf{q}}_p^{unnorm}||\nonumber
\end{gather}

\hrule
\vspace{2mm}
\textbf{Reset}
\begin{gather}\label{eq:MEKF:Reset}
    \Delta\hat{\textbf{x}}^+ = 0\\\nonumber
    \hat{\textbf{q}}^+ = \hat{\textbf{q}}_p^+
\end{gather}

\hrule
\vspace{2mm}
$\hat{\textbf{P}}$ in Eq. (\ref{eq:MEKF:PredictionStep}) deviates from \cite{b1} by not following $\textbf{FP}+\textbf{PF}^T$, which enables different sampling rates for the gyroscope and the star tracker, which is more realistic.
Note that the term $\begin{bmatrix}
        \omega\\
        0
    \end{bmatrix}
    \otimes \textbf{q}$ from Eq. (\ref{eq:quatkinematics}) is the $\Xi$ term found in both Eqs. (\ref{eq:MEKF:PredictionStep}) and (\ref{eq:MEKF:UpdateStep})
\section{Simulation}\label{Sec:simulation}

The implementation is carried out in Simulink, a programming environment based on MATLAB for modelling, simulating and analyzing dynamic systems.
It is possible to add further functionality to the base Simulink experience with the use of toolboxes. For the implementation, \textit{Robotics System Toolbox} has been used to easily convert between quaternions, rotation matrices, and Euler angles.

\subsection{Trajectory}
The simulated trajectory is made to emulate a real low Earth orbit satellite, where the average orbit time is $90$ minutes, without taking the rotation of the Earth into consideration.
The quaternion graph of an orbit like this would look sinusoidal. For this reason, it is possible to describe it with a cosine function
\begin{equation}\label{eq:realisticTrajectory}
    \omega = -\cos(\frac{1}{88\cdot 60}t\cdot 2\pi)\cdot \frac{\pi}{2}
\end{equation}

with $t$ being the time. As $2\pi$ is a full rotation, the frequency of rotation is $88\cdot 60$, the orbit time in seconds.

The quaternion is initialized and from Eq. (\ref{eq:quatkinematics}) the derivative $\Dot{\textbf{q}}$ is found and integrated to get $\textbf{q}$ which is used iteratively in the equation.

White gaussian noise is added to the $\omega$, and both the quaternion and the noisy omega is fed into both the AEKF and the MEKF filters, although the quaternion is only used to initialize each filter with the correct quaternion, while the noisy omega is used throughout.
      
\subsection{Star Tracker}
\begin{figure}[htbp]
    \centering
    \includegraphics[width = 0.2\textwidth]{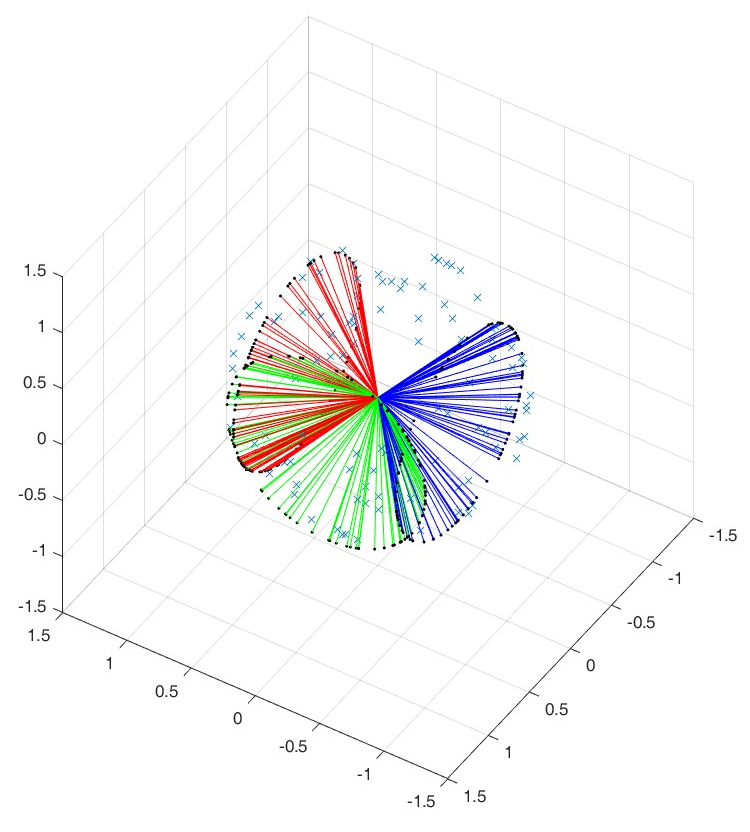}
    \caption{Three emulated star trackers observing stars in their respective field of view with the standalone cross-signs representing stars.}
    \label{fig:emulatedstartracker}
\end{figure}

For a star tracker to work, there needs to be different stars that it can detect. In the simulation, this is emulated by having $100$ stars be represented by vectors originating from the same center. It is possible to emulate $N$ stars, but with more stars, the computation time increases.
The emulated star tracker in simulink works by having a vector emerge from the focal point of the camera, and if the angle between this vector and one of the star vectors is smaller than the angle found in Eq. (\ref{eq:starangle}), then it is considered within the field of view of the camera and is used for attitude determination which outputs a quaternion that is used as the measurement in both the AEKF and the MEKF. Since the noise on a real star camera happens depending on the pixels on the lens, it is not possible to recreate that in this case. Instead, there is noise on the star vectors.
As noted in \ref{sec:startracker}, multiple star trackers provide an advantage of error minimization and redundancy, and therefore in the simulation, it is possible to have up to six star cameras that can be easily switched on and off individually. Three star trackers can be seen in Fig. \ref{fig:emulatedstartracker}.

\subsection{Comparison}
As discussed, the AEKF and the MEKF are compared in terms of certainty by finding the norm of the covariance matrix and the conditioning of the covariance matrix. Furthermore the actual differences of the estimation against the true quaternion are examined in terms of the norm of the absolute error and differences in roll, pitch, and yaw estimations. 

\section{Results}\label{Sec:results}
\begin{figure}[ht]
\centerline{\includegraphics[width = 0.4\textwidth]{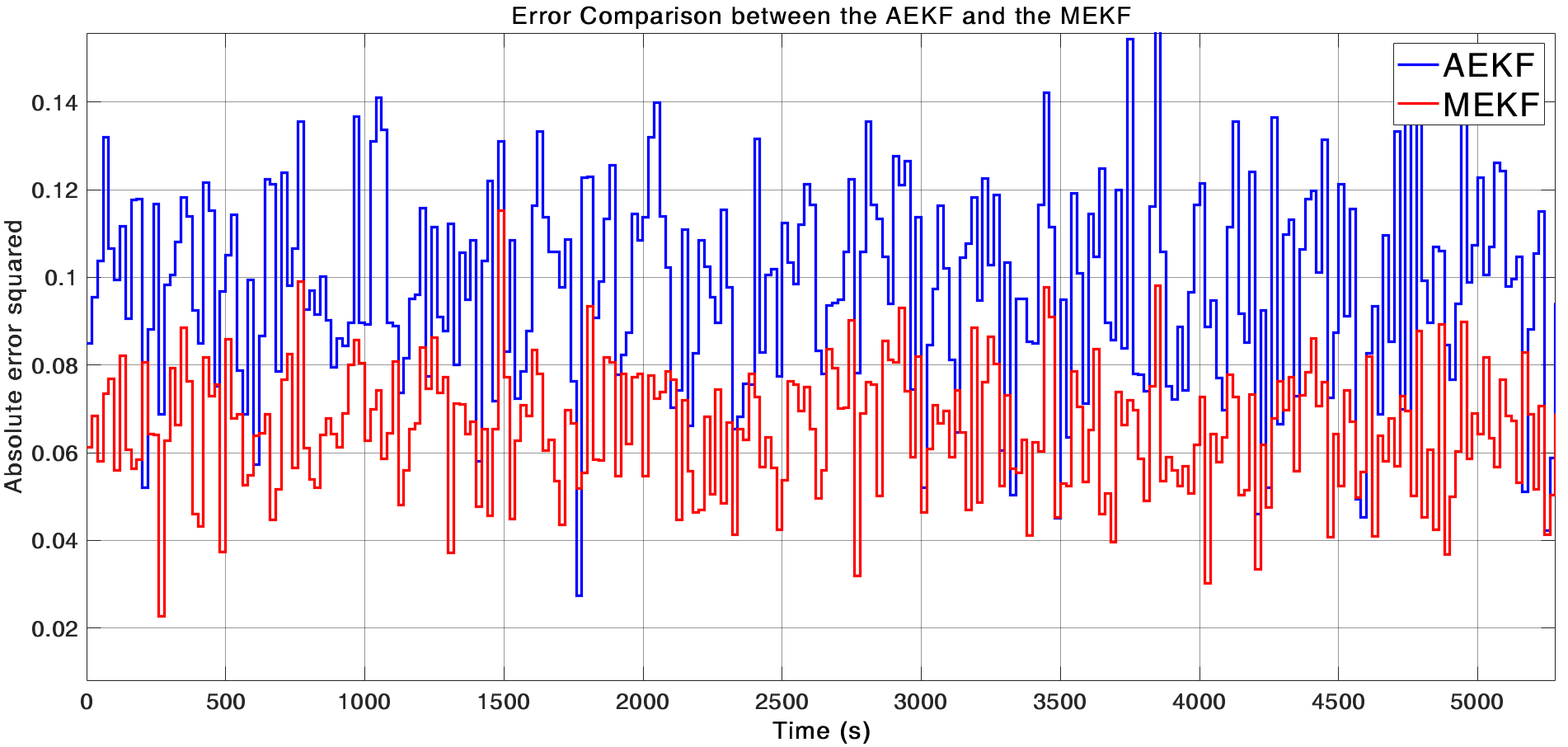}}
\caption{Absolute error comparison between the AEKF and the MEKF}
\label{fig:error1}
\end{figure}

The AEKF and MEKF have been compared in the basis of absolute error, uncertainty, computational time, and conditioning number. The first metric was the performance of the estimation algorithm and here the true quaternion was compared against the estimated quaternion as seen in Fig. \ref{fig:error1}. It can be seen that quaternion estimation is much more robust in the MEKF and has a much lower error as was expected. The MEKF has a lower error variance and a much lower bias. The AKEF has a much larger error variance and much larger bias compared to the MEKF as was expected.
The other metrics can be seen in Table \ref{table:differences}. The uncertainty is found from the norm of the covariance matrix, and it can be seen, the MEKF has a lower uncertainty, which is expected and this does align with the previous metric as MEKF has the better estimation.
The next metric, computational time, was found by taking the mean of $100$ iterations. It can be seen that the AEKF is faster than the MEKF, which is somewhat expected. While the lower dimensionality of the MEKF would suggest a faster computational time, the more complex operations in the filter itself suggests the opposite. Here, the AEKF is faster because of the relatively simple addition in comparison to the multiplication going on in the MEKF.
The last metric, the conditioning number, is $1$ and the same for both filters, which is unexpected, as the literature mentions that the AEKF would have a larger conditioning number.

\begin{table}[ht]
\caption{Different compared parameters}
\centering
\begin{tabular}{l|c|c|c}
\textbf{} & \textbf{Uncertainty} & \textbf{Computational time} & \textbf{Conditioning number} \\ \hline
\textbf{MEKF} & 0.01 & 0.012 s & 1 \\ \hline
\textbf{AEKF} & 1 & 0.002 s & 1
\end{tabular}
\label{table:differences}
\end{table}

\section{Discussion}\label{Sec:discussion}
Two different derivations of the KF, specifically the AEKF and the MEKF have been implemented in simulation with the goal of comparing the two to evaluate if the MEKF is so superior as stated in previous work. Each filter uses the quaternion representation, where the norm is constrained to $1$. Each instance of the filters uses measurements obtained from multiple emulated star trackers, where the Davenport q-method algorithm has been employed to determine the attitude of the satellite. The key difference between the two filters is in the way the error is obtained. As the names may suggest, the error quaternion is obtained through addition in AEKF and through the quaternion multiplication in MEKF.
The results from the study shows that the MEKF is more stable than the AEKF in terms of the norm of the absolute error. It was also seen that the computational time of the MEKF was worse than that of the AEKF which was half-expected. While it has a lower dimensionality, it also performs more complicated operations in comparison. This different, however, is almost negligible in a real world scenario. The conditioning number of both the MEKF and the AEKF were both equal to $1$, which differs from the literature, where it was stated that the AEKF would have a higher conditioning number. A large conditioning number indicates that the matrix being inverted is poorly conditioned, meaning that it is close to being singular or degenerate. In this case, neither filter were close to being singular.
\section{Conclusion}\label{Sec:conclusion}
The aim of the paper was to compare the Additive and the Multiplicative Extended Kalman Filter in a more practical sense than in the related works. To this end, implementations were simulated with multiple star trackers and a gyroscope as measurements to determine the attitude of a satellite.
The results were somewhat expected and showed the MEKF with a smaller absolute error than the AEKF and was more stable in its estimation. The comparison between the covariance matrices showed that the covariance matrix of the MEKF had more certainty than the AEKF, but was slower in terms of computation time. It was found that the conditioning number for both filters were equal, which means that the covariance matrix of the AEKF is not singular as stated in the literature. Overall, the MEKF had much better stability and was more resilient to noisy measurements.
The paper can conclude that based on the lower error, uncertainty and lower variance of the error, as well as the fact that the computational time is so low that the difference is almost negligible, the MEKF is better than the AEKF in the context of satellite attitude determination, which supports current opinion in the satellite attitude community.

\vspace{12pt}

\end{document}